\documentclass[pdflatex,sn-mathphys-num]{sn-jnl}

\usepackage{graphicx}%
\usepackage{multirow}%
\usepackage{amsmath,amssymb,amsfonts}%
\usepackage{amsthm}%
\usepackage{mathrsfs}%
\usepackage[title]{appendix}%
\usepackage{xcolor}%
\usepackage{textcomp}%
\usepackage{manyfoot}%
\usepackage{booktabs}%
\usepackage{algorithm}%
\usepackage{algorithmicx}%
\usepackage{algpseudocode}%
\usepackage{listings}%
\usepackage{eurosym}%

\theoremstyle{thmstyleone}%
\theoremstyle{thmstyletwo}%
\newtheorem{remark}{Remark}%

\theoremstyle{thmstylethree}%

\begin{document}
\title[PolyFormer]{Learning efficient representations of complex constraints for scalable optimization}


\author[1,2]{\fnm{Yilin} \sur{Wen}}\email{wenyl@ncepu.edu.cn}
\equalcont{These authors contributed equally to this work.}

\author[3]{\fnm{Yi} \sur{Guo}}\email{yi.guo@bit.edu.cn}
\equalcont{These authors contributed equally to this work.}

\author*[1]{\fnm{Bo} \sur{Zhao}}\email{zhaobozju@163.com}

\author*[4]{\fnm{Wei} \sur{Qi}}\email{qiw@tsinghua.edu.cn}

\author*[2]{\fnm{Zechun} \sur{Hu}}\email{zechhu@tsinghua.edu.cn}

\author[5]{\fnm{Colin} \sur{Jones}}\email{colin.jones@epfl.ch}

\author[3]{\fnm{Jian} \sur{Sun}}\email{sunjian@bit.edu.cn}

\affil[1]{\orgdiv{School of Electrical and Electronic Engineering}, \orgname{North China Electric Power University}, \orgaddress{\street{Changping District}, \city{Beijing}, \postcode{102206}, \country{China}}}

\affil[2]{\orgdiv{Department of Electrical Engineering}, \orgname{Tsinghua University}, \orgaddress{\street{Haidian District}, \city{Beijing}, \postcode{100084}, \country{China}}}

\affil[3]{\orgdiv{School of Automation}, \orgname{Beijing Institute of Technology}, \orgaddress{\street{Haidian District}, \city{Beijing}, \postcode{100081}, \country{China}}}
\affil[4]{\orgdiv{Department of Industrial Engineering}, \orgname{Tsinghua University}, \orgaddress{\street{Haidian District}, \city{Beijing}, \postcode{100084}, \country{China}}}


\affil[5]{\orgdiv{Laboratory for Systems and Control}, \orgname{École Polytechnique Fédérale de Lausanne (EPFL)}, \orgaddress{\street{Station 9}, \city{Lausanne}, \postcode{1015}, \country{Switzerland}}}



\abstract{Complex constraints often make real-world optimization computationally prohibitive at the scale and speed required for operational decision-making. Here we introduce PolyFormer, a PIML framework that learns compact polytopic representations of the geometry induced by complex constraints. PolyFormer captures constraint-induced geometry and transforms it into efficient polytopic reformulations, reducing the complexity of downstream optimization and enabling the use of off-the-shelf solvers. Neural parameterizations further enable rapid adaptation to varying operating conditions without retraining. Through evaluations across three important problems, i.e., large-scale resource aggregation, network-constrained optimization, and optimization under uncertainty, PolyFormer achieves online solver speedups of up to 6,400-fold and memory reductions of up to 99.87\%, while maintaining small feasibility and objective errors. Together, these results establish learned geometric constraint representations as an effective and scalable route to prescriptive optimization under diverse forms of constraint complexity.
}




\maketitle

\section{Introduction}\label{sec:intro}

Real-world optimization problems are governed by complex constraints that limit their scalability. Three prototypical sources of complexity are pervasive across domains: the large scale of entities involved, their interdependencies, and uncertainties. Numerous per-individual constraints arise in large-scale resource control or management \cite{CHEN2022109947, Xin2022Swarm, Liang2024Meituan}. Network interdependencies appear in communication \cite{Wu2015Optimizing}, energy \cite{millinger_diversity_2025}, and transportation \cite{YIM2011351} systems. Uncertainty-related constraints emerge in many scientific \cite{Baruch2014Communicating}, engineering \cite{chester_keeping_2020}, or financial \cite{Shen2025Convex} decisions. These characteristics make optimization-based decision-making problems high-dimensional, nonlinear, and mixed-integer, substantially complicating the search for effective solutions. Such difficulties intensify as large-scale socio-technical systems continue to expand. For example, electric vehicles, whose planning and operation involve decision-making across power systems, transportation networks, and communication infrastructures, have increased by over eightyfold from 2014 to 2024 \cite{IEA_GlobalEVOutlook2025}, dramatically increasing the dimensionality of decision spaces. The Iberian blackout of 28 April 2025 affected around 50 million people and caused estimated economic losses of up to EUR~3.5 billion, illustrating the wide societal and economic consequences of large-scale power-system disruptions \cite{Batlle2025blackout,ENTSOE2026blackout}. Such events underscore the need for rapid, computationally tractable scheduling decisions that preserve voltage-control capability and operational security margins under complex network constraints.

Existing studies follow two main paradigms to address this challenge, but both face fundamental limitations.
The first paradigm comprises analytical methods. These include decomposition and coordination that split large problems into smaller subproblems solved iteratively \cite{adulyasak2015benders, MAL-016, Dai2025Advancing}, convex relaxation approaches that approximate nonconvex constraints with tractable convex ones \cite{Venkat2013Computational, Low2014Convex, bemporad_algorithm_2006}, and analytical surrogate models that simplify complex feasible regions \cite{jones_polytopic_2010, wen_aggregate_2022, Wu2024Data}. While these methods provide theoretical guarantees, they often require problem-specific derivations, potentially suffer from unstable convergence, and may lack scalability across diverse constraint types.
The second is based on machine learning (ML). Some approaches directly predict optimization solutions \cite{Donti_2021_DC3, wei_deep_2025}. Others aim to accelerate solvers by learning branch-and-bound policies \cite{he_learning_2014}, generating warm starts \cite{baker2019learning}, or improving heuristic search \cite{shen_free-energy_2025}. Prediction-based methods often lack feasibility guarantees and interpretability, and may also face training instability when applied to large-scale problems \cite{WANG2025100057}. Solver-acceleration methods still require storing and operating on the full optimization model, and their worst-case solution times remain prohibitively long. These issues limit their use in real-time or memory-constrained settings.

In this context, physics-informed machine learning (PIML) has emerged as a natural integration of machine learning and the analytical modelling of physical laws or prior knowledge. By embedding physical principles into the learning framework, PIML can improve data efficiency and promote the physical consistency of the results \cite{karniadakis_physics-informed_2021}. This approach has led to advancements in various fields, including physics \cite{Maziar2020Hidden,rao_encoding_2023}, biology \cite{feng_sliding-attention_2024}, chemistry \cite{venturella_unified_2025}, geosciences \cite{singh2024piml}, and others. However, a closer examination of the literature reveals a significant gap: existing research has primarily focused on \emph{predictive} tasks, where models are trained to satisfy differential equations or other physical constraints by minimizing residual errors. In contrast, \emph{prescriptive} tasks involving high-dimensional or topologically complex constraints have not been fully explored within the PIML framework. In these tasks, current PIML architectures struggle to guarantee the reliability needed for valid solutions. Only a few attempts have been made to use physics-informed neural networks as surrogate models \cite{SENHORA2022115116} or for direct solution prediction \cite{PATEL2024775}, but these approaches may conflict with optimization objectives and are difficult to generalize across different domains. This highlights a key question: while PIML has been widely used for predictive modelling, can it also serve prescriptive optimization tasks under complex physical constraints? Unlocking this capability requires a carefully designed adaptation of PIML that explicitly balances feasibility and optimality in prescriptive optimization.

\begin{figure}[htbp]
\centering
\includegraphics[width=0.93\textwidth]{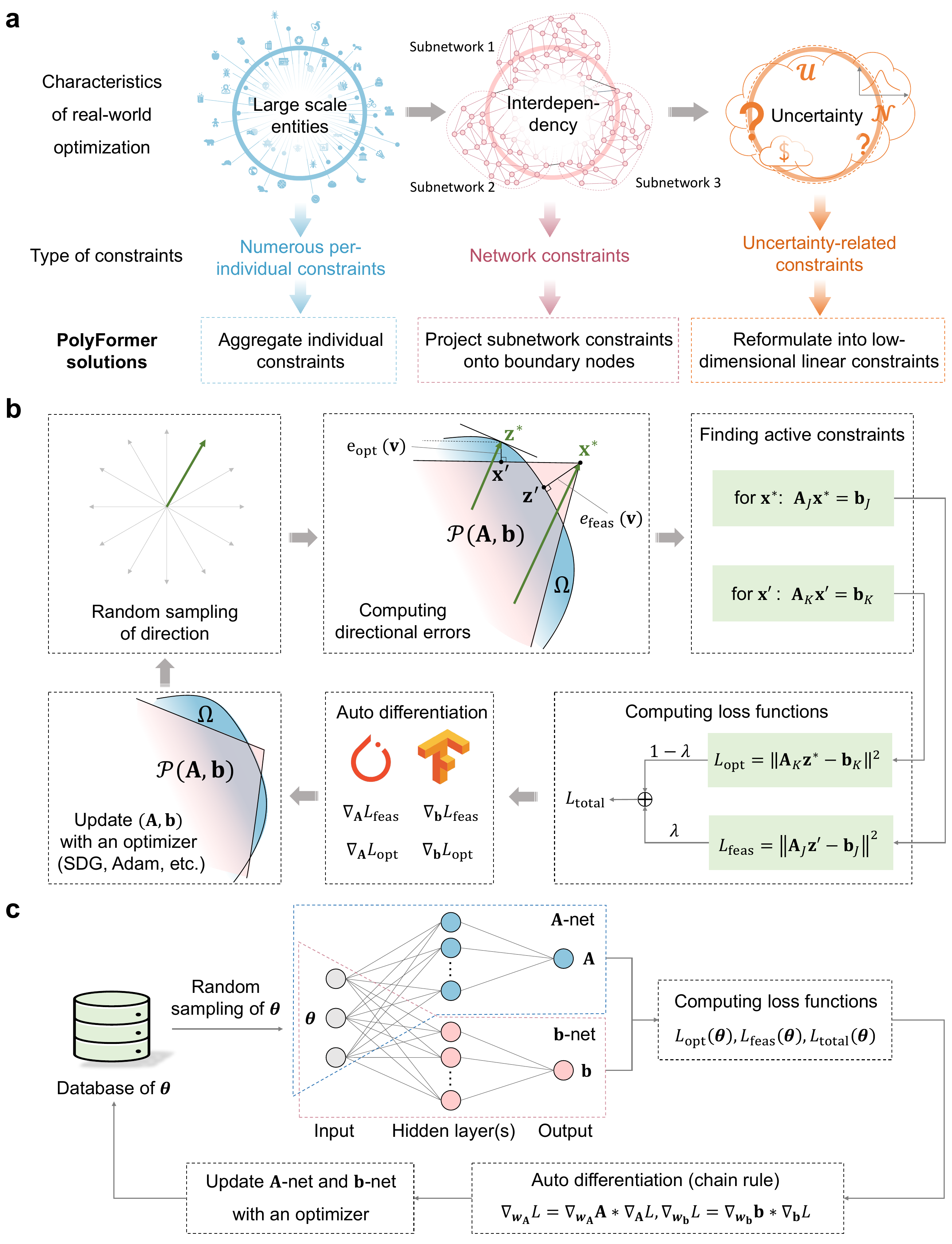}
\caption{\textbf{Capability and training procedure of PolyFormer.} 
\textbf{a,} PolyFormer solutions under three classes of complex constraints: large-scale per-individual constraints, network constraints, and uncertainty-related constraints. 
\textbf{b,} Training pipeline of PolyFormer. The procedure comprises six steps: (1) random sampling of a direction $\mathbf{v}$; (2) computation of the directional feasibility and optimality errors, $e_{\mathrm{feas}}(\mathbf{v})$ and $e_{\mathrm{opt}}(\mathbf{v})$, along with the associated boundary points $\mathbf{x}^{\star}$, $\mathbf{z}'$, $\mathbf{z}^{\star}$ and $\mathbf{x}'$ (see ``Error metrics'' section in the Methods); (3) identification of active constraints in the current polytope at $\mathbf{x}^{\star}$ and $\mathbf{x}'$; (4) evaluation of the loss based on the distances from $\mathbf{z}'$ and $\mathbf{z}^{\star}$ to the hyperplanes defined by the corresponding active constraints; (5) gradient computation via automatic differentiation (see the ``Loss function and gradients'' section in the Methods); and (6) parameter updates of $\mathbf{A}$ and $\mathbf{b}$ using gradient-based optimization. 
\textbf{c,} Training of the parameterized PolyFormer additionally involves sampling a parameter $\boldsymbol{\theta}$ and mapping it to $\mathbf{A}(\boldsymbol{\theta})$ and $\mathbf{b}(\boldsymbol{\theta})$ using two neural networks, denoted $\mathbf{A}$-net and $\mathbf{b}$-net, with trainable weights $\boldsymbol{w}_{\mathbf{A}}$ and $\boldsymbol{w}_{\mathbf{b}}$, respectively.}
\label{fig:framework}
\end{figure}

Here, we present a new capability of PIML. Instead of using it to predict solutions or accelerate existing solvers, we introduce \textbf{PolyFormer}, a PIML framework that learns to \textit{simplify the problem itself}. The key insight is that many complex constraints, despite their diverse forms, define a geometric shape that can be learned by PIML and approximated by a compact polytopic reformulation. By replacing the original constraints with their simplified polytopic representations, PolyFormer enables more efficient solution by off-the-shelf optimization solvers, with tunable prioritization of feasibility over optimality. 
PolyFormer is a targeted design of PIML to optimization with three notable features:
(1) it directly handles mixed continuous and discrete variables, unlike many analytical approximation methods restricted to continuous-variable formulations; 
(2) it controls the trade-off between feasibility and optimality, allowing adaptation to different application requirements; 
and (3) it adapts to varying parameters through neural network parameterization, enabling fast re-estimation without retraining. These design features enable PolyFormer to address the three aforementioned prototypical challenges that limit scalability in optimization under complex constraints (see Fig. \ref{fig:framework}a). It should be emphasized that PolyFormer differs fundamentally from existing constraint learning approaches \cite{FAJEMISIN20241, maragno_mixed-integer_2025}. Whereas the latter approaches learn constraints from data, PolyFormer explicitly incorporates known mechanistic or mathematical constraint models, including physics-based models. This significantly reduces the dependence on large-scale training data, while also enabling the framework to adapt to parameter variations in physical models, which is common in real-world applications.

We evaluate the performance of PolyFormer on three important problems that are representative of the three challenges in constrained optimization. 
First, to address the numerous per-individual constraints arising from the large-scale entities, PolyFormer learns to aggregate them to produce a compact set of constraints defined over aggregated variables. PolyFormer removes 99.83\% of constraints in the 1,000-resource case and 98.47\% in the mixed-variable case while maintaining high approximation fidelity. Unlike many existing analytical methods, it directly processes models containing mixed continuous and discrete auxiliary variables.
Second, regarding the complex interdependencies among the entities, which are typically characterized by large-scale network constraints, PolyFormer projects subnetwork constraints onto boundary nodes, producing a polytopic feasible set defined solely over boundary node variables. Experimental results demonstrate an online solver speedup of up to 6,400$\times$ and a 99.6\% reduction in memory usage, while keeping the maximum feasibility and objective errors on the order of \(10^{-8}\) and \(10^{-4}\), respectively.
Third, for optimization under uncertainty-related constraints, whose standard formulations introduce substantial auxiliary variables, PolyFormer generates low-dimensional linear reformulations that reduce problem size from up to 1,044,497 constraints to 1,617. The resulting strategies attain competitive risk-return trade-offs relative to DRCC-linear.
Collectively, these results demonstrate that PolyFormer transforms diverse complex constraints into compact, solver-compatible representations, establishing learned geometric simplification as a scalable paradigm for prescriptive optimization.

\section{Results}
\subsection{Overview of PolyFormer}

Consider the region $\Omega$ as a nonempty, compact subset of the $n$-dimensional Euclidean space $\mathbb{R}^n$. Let $\mathcal X\subseteq\mathbb R^n$ and $\mathcal Y\subseteq\mathbb R^{n'}$ denote the domains of the primary and auxiliary variables, respectively. The region is generally formulated as
\begin{equation}\label{eq:orig_region}
    \Omega  \triangleq \left\{\mathbf{x}\in\mathcal X\,\middle|\,\exists\,\mathbf{y}\in\mathcal Y:\ {g_j}({\mathbf{x}},{\mathbf{y}}) \le 0,\ j = 1,2, \ldots, m\right\}.
\end{equation}
Here, $\mathbf{x}$ contains the variables of primary interest, whereas $\mathbf{y}$ contains auxiliary variables. When $n'=0$, the problem reduces to fitting a set directly in $\mathbb R^n$. When $n'>0$, $\Omega$ is the projection of a feasible set in $\mathbb R^{n+n'}$ onto the $\mathbf{x}$-space. Both settings can be processed by the PolyFormer pipeline. The integer $m$ denotes the number of original constraints, and the inequalities $g_j(\mathbf{x},\mathbf{y})\leq0$ encode the mechanistic, mathematical, or operational constraints that determine the geometry of $\Omega$.

PolyFormer approximates $\Omega$ using a polytope $\mathcal P(\mathbf A,\mathbf b)=\left\{\mathbf x\mid\mathbf A\mathbf x\leq\mathbf b\right\}$, where $\mathbf{A}$ is an $M \times n$ matrix and $\mathbf{b}$ is an $M$-dimensional vector. Here, $M$ represents the number of hyperplanes that define the polytope ${\mathcal P}$, which is a predetermined hyperparameter constrained by practical factors such as memory and computational time. If $\Omega$ is convex, then ${\mathcal P}$ can approximate $\Omega$ to arbitrary precision provided that $M$ is sufficiently large \cite{aryarahul_optimal_2022}. When $\Omega$ is nonconvex or disconnected, the current PolyFormer remains implementable, but the resulting polytope may approximate the convex hull of $\Omega$ rather than its exact geometry (see ``Error metrics'' and ``Typical geometries'' sections in the Methods). 

The foundation of fitting is defining an appropriate discrepancy measure between the polytope ${\mathcal P}({\mathbf{A}},{\mathbf{b}})$ and the original region $\Omega$. Motivated by practical optimization requirements, we use two complementary concepts: feasibility error and optimality error. The feasibility error quantifies overestimation of $\Omega$ by ${\mathcal P}$, reflecting the risk that ${\mathcal P}$ admits decisions that violate the original constraints. The optimality error quantifies undercoverage of $\Omega$ by ${\mathcal P}$, reflecting the risk that ${\mathcal P}$ excludes solutions attainable in the original region. To compute these concepts tractably, we define directional counterparts using support points along each direction $\mathbf v$. The directional feasibility error $e_{\mathrm{feas}}(\mathbf v)$ is the squared distance from the support point of ${\mathcal P}$ to $\Omega$. Conversely, $e_{\mathrm{opt}}(\mathbf v)$ is the squared distance from the support point of $\Omega$ to ${\mathcal P}$. For compact convex sets, $e_{\mathrm{feas}}(\mathbf v)=0$ for all directions if and only if ${\mathcal P}\subseteq\Omega$. Likewise, $e_{\mathrm{opt}}(\mathbf v)=0$ for all directions if and only if $\Omega\subseteq{\mathcal P}$. Therefore, jointly reducing the two errors balances feasibility and optimality in the polytopic approximation of $\Omega$. For nonconvex $\Omega$, $e_{\mathrm{opt}}(\mathbf v)$ measures discrepancies between ${\mathcal P}$ and $\operatorname{conv}(\Omega)$ because $\Omega$ and its convex hull share the same support function. Meanwhile, $e_{\mathrm{feas}}(\mathbf v)$ detects nonconvex boundary violations encountered through projection onto $\Omega$. Together, the two errors balance convex-hull coverage against these boundary violations. We define the aggregate errors $e_{\mathrm{feas}}$ and $e_{\mathrm{opt}}$ as expectations of their directional counterparts over an isotropic distribution of directions (see ``Error metrics'' section in the Methods).

PolyFormer identifies a polytope ${{\mathcal P}}({\mathbf{A}},{\mathbf{b}})$ that best fits $\Omega$ by minimizing a weighted sum of the feasibility and optimality errors, as formulated by \eqref{eq:res:min_weighted_hausdorff}.
\begin{equation}\label{eq:res:min_weighted_hausdorff}
({\mathbf{A}^\star},{\mathbf{b}^\star}) = \mathop {\arg \min }\limits_{{\mathbf{A}\in{\mathbb{R}^{M\times n}}},{\mathbf{b}\in{\mathbb{R}^{M}}}} {\lambda {e_{{\text{feas}}}} + (1 - \lambda ){e_{{\text{opt}}}}},
\end{equation}
where $\lambda \in [0, 1]$ controls the trade-off between the feasibility and optimality errors. By adjusting $\lambda$, the learned polytope targets an inner approximation ($\lambda=1$ prioritizes feasibility), an outer approximation ($\lambda=0$ prioritizes coverage), or an intermediate trade-off ($0<\lambda<1$) to better fit the shape of the feasible region. This flexibility allows the two errors to be balanced according to the application context, as demonstrated in the following three subsections.

The feasibility and optimality errors do not admit explicit gradients with respect to $\mathbf A$ and $\mathbf b$ because they depend on solutions to nested support and projection problems. PolyFormer addresses this bottleneck by converting these geometric approximation errors into an explicit surrogate loss that is differentiable at each iteration (see ``Loss function and gradients'' section in the Methods). The surrogate and its gradients are derived from geometric analyses of the discrepancy between $\Omega$ and its polytopic approximation $\mathcal P$. This differs from most existing PIML approaches, in which physical losses are typically constructed from residuals of governing equations \cite{Maziar2020Hidden,rao_encoding_2023,feng_sliding-attention_2024,venturella_unified_2025,singh2024piml,SENHORA2022115116,PATEL2024775}. The differentiable surrogate enables efficient gradient computation using frameworks such as PyTorch \cite{paszke_pytorch_2019} or TensorFlow \cite{abadi2016tensorflow}. Gradient-based optimizers, including stochastic gradient descent (SGD) \cite{rumelhart_learning_1986} and Adam \cite{Diederik2015Adam}, then update $\mathbf A$ and $\mathbf b$. These components provide a practical training procedure for addressing \eqref{eq:res:min_weighted_hausdorff}, as illustrated in Fig. \ref{fig:framework}b. When $\lambda\in(0,1)$, the two objectives dynamically balance feasibility and optimality during fitting.

The training procedure described above enables PolyFormer to fit a fixed target region $\Omega$. However, many practical scenarios require repeatedly estimating regions that change with varying parameters, motivating the extension of PolyFormer to handle parameterized regions $\Omega(\boldsymbol{\theta})$ (where $\boldsymbol{\theta}$ denotes the varying parameters). Such parameters arise naturally in real-world applications. Examples include environmental conditions such as temperature, humidity, and pressure in control systems, as well as socioeconomic conditions such as investment levels and population density in public-service systems. To accommodate these scenarios, PolyFormer treats $\mathbf{A}$ and $\mathbf{b}$ as functions of $\boldsymbol{\theta}$, i.e., $\mathbf{A}(\boldsymbol{\theta})$ and $\mathbf{b}(\boldsymbol{\theta})$, which are fitted using neural networks, a well-established paradigm for solving high-dimensional fitting problems \cite{rumelhart_learning_1986}. Training these neural networks simply adds a sampling step over $\boldsymbol{\theta}$ to the aforementioned procedure, as illustrated in Fig.~\ref{fig:framework}c. By incorporating varying parameters, PolyFormer attains greater adaptivity, enabling rapid re-estimation of the region following parameter updates and thereby substantially reducing computational costs in large-scale applications.

During PolyFormer training, the initialization, normalization, and selection of the weight coefficient $\lambda$ influence model performance. We outline our experimental strategies for these aspects in ``Implementation remarks'' section in the Methods to facilitate smooth deployment of PolyFormer in practice. To further clarify PolyFormer's training process, its adaptation to changes in $\boldsymbol{\theta}$, and its scalability across dimensions, we evaluate PolyFormer on five representative geometries, including 2D polygons, ellipses, and nonconvex regions, as well as $2$ to $200$-dimensional hypercubes and balls (see ``Typical geometries'' section in the Methods). We also provide a user guide in \textcolor{blue}{Supplementary Note~1}, which includes a detailed description of the PolyFormer workflow and a step-by-step example to help readers quickly get started. Collectively, these practical guidelines, illustrative geometries, and hands-on instructions form a coherent toolkit that links the methodological design of PolyFormer with its effective use in practice, positioning PolyFormer as a ready-to-use framework for simplifying complex constraints and supporting scalable optimization across scientific, engineering, and socio-economic applications.

\subsection{Aggregation of large-scale resources}

The first challenge of solving real-world optimization problems that PolyFormer addresses is the complexity of numerous per-individual constraints, where each entity is governed by its own physical model and operational constraints. In such cases, aggregation techniques, i.e., representing a large number of entities using aggregated variables and constraints, are commonly employed to reduce model complexity \cite{Rogers1991Aggregation}. Typical examples include logistics and delivery scheduling, where multiple orders are aggregated to enable efficient task assignment \cite{Liang2024Meituan}; cloud computing resource management, where geographically proximate computing units are aggregated to support coordinated allocation \cite{khezri_dljsf_2024}; and large-scale control systems, where the control capabilities of individual devices are aggregated to define the boundaries for higher-level control signals \cite{Fattahi2025Utilities,chen_aggregate_2020, wang_aggregate_2021}. Mathematically, aggregation corresponds to computing the Minkowski sum of multiple sets, which is generally NP-hard \cite{tiwary_hardness_2008}. PolyFormer offers a tractable solution for approximating the aggregation, substantially reducing the complexity of handling large-scale, heterogeneous resources.

\begin{figure}[htbp]
\centering
\includegraphics[width=0.95\textwidth]{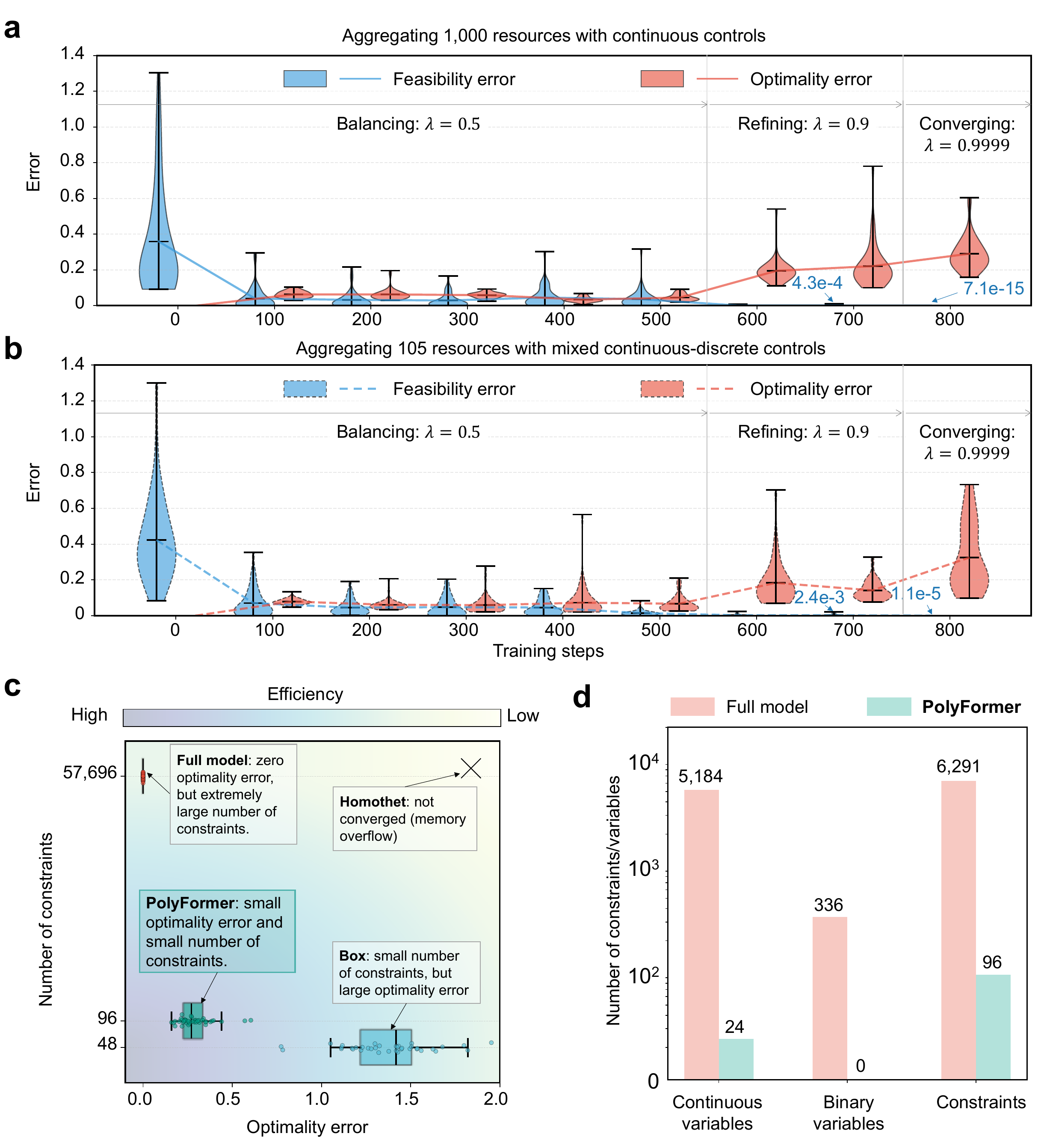}
\caption{\textbf{Training process and benchmark results for large-scale resource aggregation. a--b,} Evolution of the feasibility and optimality error distributions during PolyFormer training for two scenarios: (a) 1,000 resources with continuous controls and (b) 105 resources with mixed continuous-discrete controls. At each displayed iteration, the error distributions are evaluated over 50 randomly sampled directions using \eqref{eq:dir_err_feas} and \eqref{eq:dir_err_opt}. Training proceeds in three phases. In the balancing phase (iterations 1--500, \(\lambda = 0.5\)), both errors decrease as they compete with each other, and the polytope adapts to the true aggregated region. The refining phase (iterations 501--700, \(\lambda = 0.9\)) prioritizes feasibility while considering optimality to a lesser extent, reducing the average feasibility error to \(4.3\times10^{-4}\) and \(2.4\times10^{-3}\), respectively. The converging phase (iterations 701--800, \(\lambda = 0.9999\)) reduces the mean feasibility errors to \(7.1\times10^{-15}\) and \(1.1\times10^{-5}\), respectively. The total training time is 124 min (a) and 65 min (b). \textbf{c,} Modelling efficiency comparison among the full model, Box, Homothet, and PolyFormer in scenario (a), plotting optimality error (conservatism) versus number of constraints (complexity). The box plots show the distribution of optimality errors. Background shading from purple to yellow indicates decreasing efficiency in balancing complexity and conservatism, with the lower-left region most desirable. PolyFormer provides the most favourable complexity--conservatism trade-off among the evaluated solutions. \textbf{d,} Complexity reduction in scenario (b): PolyFormer removes all 336 binary variables and reduces continuous variables from 5,184 to 24 and constraints from 6,291 to 96.}\label{fig:aggregation}
\end{figure}

We consider the aggregation of feasible operational regions for controllable devices, including electric vehicles (EVs), battery storage systems (BSSs), and heat pumps (HPs). This is a highly active topic in smart grid applications for maintaining dynamic electricity supply-demand balance amid the proliferation of intermittent renewable energy throughout the modern energy system \cite{chen_aggregate_2020,wang_aggregate_2021}. A tight inner approximation is essential here because overestimating aggregate flexibility can produce infeasible dispatch instructions. We examine two experimental scenarios: the first aggregates 1,000 resources with continuous control inputs; the second aggregates 105 resources with mixed continuous-discrete controls. The formal problem definition and descriptions of device models are provided in the ``Apply PolyFormer to resource aggregation'' section in the Methods, with detailed formulations and numerical settings in \textcolor{blue}{Supplementary Note 3}. Figs. \ref{fig:aggregation}a and \ref{fig:aggregation}b illustrate the evolution of feasibility and optimality errors during training in the two scenarios, respectively, showing that PolyFormer converges to near-inner approximations with mean feasibility errors of \(7.1\times10^{-15}\) and \(1.1\times10^{-5}\), respectively.

To assess the quality of the approximation produced by PolyFormer, we benchmark it against the full model of all resources and two state-of-the-art aggregation techniques, namely the Box \cite{chen_aggregate_2020} and Homothet \cite{wang_aggregate_2021} methods. Fig. \ref{fig:aggregation}c reports the number of constraints (reflecting model complexity) and the optimality errors (reflecting conservatism) of different methods under the 1,000-resource case. Feasibility errors for all methods are below $10^{-5}$ and are therefore omitted. Although the full model is theoretically exact (with zero feasibility and optimality errors), it requires an extremely large number of constraints (57,696 constraints). The Homothet method fails to converge due to memory overflow, yielding no valid solution. The Box method has the minimum model complexity (only 48 constraints), but its optimality error is the highest, indicating a poor fit to the true aggregated region. In contrast, PolyFormer exhibits a small optimality error (approximately 21\% of that of the Box method) while maintaining a small number of constraints (96 constraints, only 0.17\% of the full model). This makes PolyFormer a compact yet accurate representation of the complex original region of all devices in the high-dimensional space.

In the second scenario, which involves discrete controls, the Box and Homothet methods are inapplicable in their original forms because of their structural assumptions. PolyFormer nevertheless processes the mixed continuous-discrete auxiliary model while attaining optimality errors comparable to those in the continuous scenario. This result demonstrates a practical mixed-variable aggregation capability beyond the scope of the two analytical baselines. As shown in Fig.~\ref{fig:aggregation}d, PolyFormer also significantly reduces model complexity relative to the full formulation: all binary variables are eliminated, the number of continuous variables is reduced by 99.54\%, and the number of constraints decreases by 98.47\%, demonstrating a substantial simplification of the original model.

In summary, PolyFormer enables a substantial reduction in model size when aggregating large populations of heterogeneous resources, while maintaining a near-inner approximation of the feasible region. This addresses the first challenge of scalable optimization, i.e., the large number of entities involved. In applications that require efficient management and control of large-scale resources, such as logistics networks \cite{Liang2024Meituan}, cloud computing infrastructures \cite{khezri_dljsf_2024}, large-scale control systems \cite{Fattahi2025Utilities}, and smart grids \cite{chen_aggregate_2020, wang_aggregate_2021}, the efficient aggregation capability provided by PolyFormer offers broad practical potential.

\subsection{Network-constrained optimization}

We next apply PolyFormer to address the second type of complex constraints illustrated in Fig. \ref{fig:framework}a, i.e., network constraints that capture the interdependency among many entities. Optimization under large-scale network constraints is common in communications \cite{Wu2015Optimizing}, energy \cite{millinger_diversity_2025}, and transportation systems \cite{YIM2011351}, and is often computationally expensive. Large-scale networks are often decomposed into smaller subnetworks. For each subnetwork, PolyFormer learns a simplified polytopic constraint defined only over its boundary nodes. This polytopic set serves as a surrogate for the original subnetwork constraints. By replacing all subnetworks with their compact boundary representations, the overall problem size is significantly reduced, leading to substantial computational savings.

\begin{figure}[htbp]
\centering
\includegraphics[width=1\textwidth]{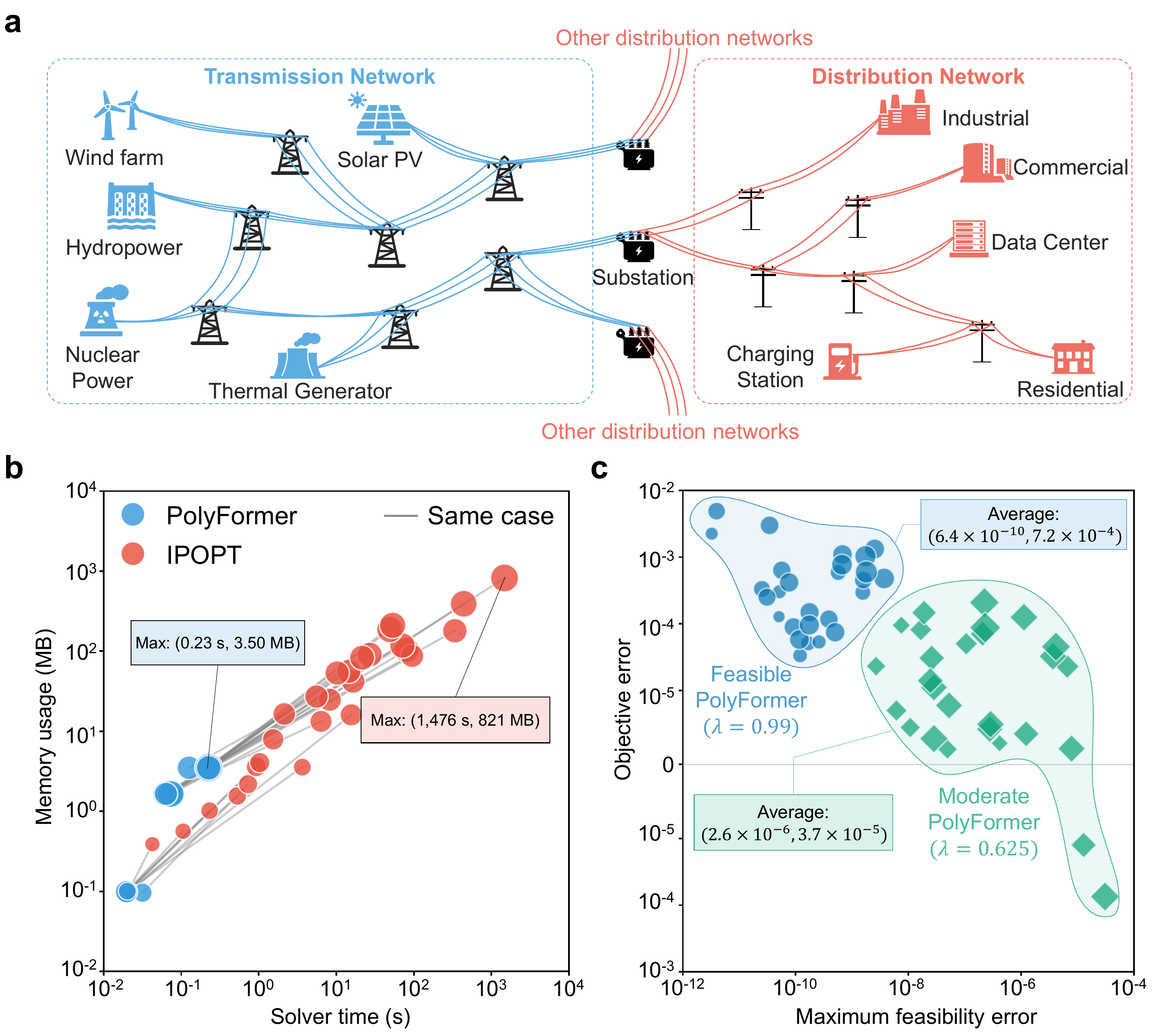}
\caption{
    \textbf{Benchmark results of the two-layer power system optimization problem. a,} Schematic diagram of the two-layer power system. The upper level transmission network includes generation units powered by various energy sources, transmitting electricity to substations. Then the distribution network distributes electricity to various users. \textbf{b,} Comparison of solver time and peak memory usage between optimization using the simplified PolyFormer model and optimization of the original full model solved by IPOPT. Each data point represents a transmission-distribution system case, with the size of the point indicating the case's complexity (the total number of transmission and distribution nodes). Each red-blue pair represents the same case, with the connection from the upper-right red points to the lower-left blue points illustrating that PolyFormer consistently reduces both computation time and memory usage compared to directly solving the full problem with IPOPT. The largest case is also annotated in the figure: solving the original model takes 1,476 seconds and 821 MB of memory, while the PolyFormer-simplified model only requires 0.23 seconds and 3.50 MB of memory. \textbf{c,} Maximum feasibility error and objective error of two PolyFormer variants. The objective error is measured relative to the objective value returned by IPOPT for the original full model, which serves as the computational reference. All feasibility error values are normalized by dividing by the sum of squares of the baseline active and reactive powers of each distribution system. ``Moderate PolyFormer'' ($\lambda = 0.625$) yields an average maximum feasibility error of $2.6 \times 10^{-6}$ and an average objective error of $3.7 \times 10^{-5}$. Two points show negative objective errors, indicating that the PolyFormer-simplified model yields lower costs than the original model. These points also have larger feasibility errors. ``Feasible PolyFormer'' ($\lambda = 0.99$) has all of the maximum feasibility errors below $10^{-8}$, with an average of $6.4 \times 10^{-10}$. However, the average objective error is $7.2 \times 10^{-4}$, larger than that of Moderate PolyFormer, indicating greater conservatism, as expected.
}\label{fig:td_case}
\end{figure}

We consider the optimal power flow \cite{donon2024gnnopf, Burchett1982Large, Dai2025Advancing}, a fundamental problem in electric power system operation, on a two-layer network comprising an upper-level transmission network and multiple lower-level distribution networks (Fig. \ref{fig:td_case}a).
The decision variables include the generation power in the transmission network, as well as the flexible power of nodes in the distribution networks. The optimization objective is to minimize the system's total operating cost. Both levels are governed by nonlinear power flow equations due to the physical laws of electricity, and engineering requirements for network safety (see \textcolor{blue}{Supplementary Note 4} for detailed formulations). An efficient solution to this problem has attracted increasing attention as the growing penetration of distributed renewable energy and the rapid expansion of large-scale computing facilities have significantly increased the interdependence and operational complexity of modern power systems \cite{colangelo_ai_2025}. Using PolyFormer, the thousands of nonlinear network constraints within each distribution network are compressed into tens of inequalities that are linear in the interface active and reactive powers for a given root-node voltage. For each T-D interface, the polytope is instantiated at an initial voltage obtained from baseline power-flow calculations and subsequently embedded as fixed linear inequalities in the upper-level optimization. The full distribution-network model is then replaced by this compact boundary representation, so that the overall optimization problem scales only with the size of the transmission network. This reduction in constraint dimensionality leads to significant improvement in computational efficiency (see ``Apply PolyFormer to the two-layer power system optimization'' section in the Methods).

The transmission and distribution network data are primarily obtained from the IEEE standard database \cite{matpower}, supplemented by the data from a real-world distribution system in Zhejiang Province, China. Together, these datasets form 27 integrated transmission--distribution system cases (see \textcolor{blue}{Supplementary Note 4} for details). We compared PolyFormer with the original full transmission-distribution model solved by IPOPT \cite{wachter_implementation_2006}, treating the resulting solutions as computational references. Fig. \ref{fig:td_case}b compares the online solver time and peak memory usage of PolyFormer after offline training with those of full-network optimization by IPOPT. PolyFormer delivers orders-of-magnitude improvements in computational efficiency across all test cases. For the largest system, the solver time is over 6,400$\times$ faster, and memory usage decreases by 99.6\%. This is achieved by simplifying the original model with 715,055 constraints and 477,691 variables into an approximate model containing only 2,239 constraints and 1,785 variables. 

Fig. \ref{fig:td_case}c reports the maximum feasibility error and the objective error under two different weight parameter settings. The feasibility error is defined as the largest deviation between the operating point (active and reactive power) at the root node obtained from the PolyFormer-simplified model and the nearest operating point that satisfies the original network constraints across all distribution networks. The objective error measures the relative difference in objective values between the PolyFormer-simplified optimization and the full network optimization. When the weight coefficient \(\lambda\) is set to 0.625, the average feasibility error is \(2.6 \times 10^{-6}\), with a maximum below \(10^{-4}\). Increasing \(\lambda\) to 0.99 further improves feasibility: all feasibility errors fall below \(10^{-8}\), with an average of \(6.4 \times 10^{-10}\), representing a reduction of approximately four orders of magnitude relative to the \(\lambda = 0.625\) setting. This gain in feasibility is accompanied by a larger objective deviation relative to the IPOPT reference. The average objective error increases from \(3.7 \times 10^{-5}\) to \(7.2 \times 10^{-4}\), representing roughly a one-order-of-magnitude increase. Overall, these results quantify a clear feasibility-optimality trade-off controlled by \(\lambda\): increasing \(\lambda\) reduces constraint violations by approximately four orders of magnitude while increasing the objective error by approximately one order of magnitude. Such a trade-off is attractive when feasibility is prioritized over close agreement with the reference objective \cite{STERN201939}.

These results demonstrate that PolyFormer makes large network-constrained optimization substantially faster and more memory-efficient while preserving high-quality solutions relative to the IPOPT references. By compressing each subnetwork into a compact boundary representation, PolyFormer offers a general strategy for scaling interconnected optimization in energy, communications, and transportation systems \cite{Wu2015Optimizing,millinger_diversity_2025,YIM2011351}.

\subsection{Optimization under uncertainty}

Finally, we leverage PolyFormer to address the third type of complex constraints shown in Fig. \ref{fig:framework}a, namely uncertainty-related constraints. Accounting for uncertainty is important for optimization problems in many applications such as finance \cite{Chen2024Data}, transportation \cite{WANG2024102923}, control \cite{Coulson2022Distributionally}, supply chain management \cite{GAO2024106803}, machine learning \cite{LIN2024106755}, and others. Optimization under uncertainty aims to find robust policies that hedge against adversarial uncertainty realizations while allowing the decision-maker to tune the level of conservatism (i.e., the trade-off between nominal performance and risk protection). 
 A recently popular framework is data-driven distributionally robust chance constraints (DRCCs), which infer an uncertainty distribution from finite samples and make decisions feasible with high probability even when the inferred distribution is misspecified \cite{WANG2024102923, Coulson2022Distributionally, GAO2024106803, LIN2024106755, Chen2024Data, xie_distributionally_2021,ordoudis_energy_2021}. Despite the practical appeal of DRCCs, they encounter considerable computational burdens \cite{SHANG2022176, ho-nguyen_distributionally_2022}. Although many DRCCs admit reformulations into linear (or conic-representable) constraints, these reformulations commonly depend on a large collection of auxiliary variables and constraints \cite{xie_distributionally_2021,ordoudis_energy_2021}, leading to substantial computational cost in large-scale problems. By contrast, PolyFormer yields a compact reformulation: it replaces the feasible region defined by each group-level DRCC and its associated investment cap with a small set of constraints expressed solely in terms of the original decision variables, thereby substantially reducing model complexity and improving computational efficiency.

The first two applications use physics-based constraint models. The portfolio case examines whether the same model-informed geometric construction extends to explicitly specified non-physical constraints, with the DRCC formulation serving as the structural prior. Portfolio optimization is a fundamental topic in finance and is widely applied in the management of large-scale assets \cite{fabozzi2002legacy}. It seeks a portfolio allocation that maximizes the expected total return, subject to a prescribed risk tolerance, under the uncertain return of each asset \cite{Chen2024Data, mohajerin_esfahani_data-driven_2018, Xie2020Bicriteria}.

In our formulation, assets are partitioned into groups according to their risk levels. For each group, we impose a minimum return requirement under uncertainty, modelled as a distributionally robust chance constraint (DRCC): the probability that the group's aggregate return falls below a prescribed threshold must not exceed a given risk level for all distributions in an ambiguity set constructed from data. Because multiple asset groups are considered, the overall portfolio problem contains multiple DRCCs coupled through the total investment constraint, leading to a computationally challenging formulation. PolyFormer is applied to approximate the joint feasible region defined by each group-level DRCC and its associated investment cap, and the resulting surrogate is embedded into the portfolio optimization problem (see ``Apply PolyFormer to DRCC portfolio optimization'' section in the Methods).

\begin{figure}[htbp]
\centering
\includegraphics[width=1\textwidth]{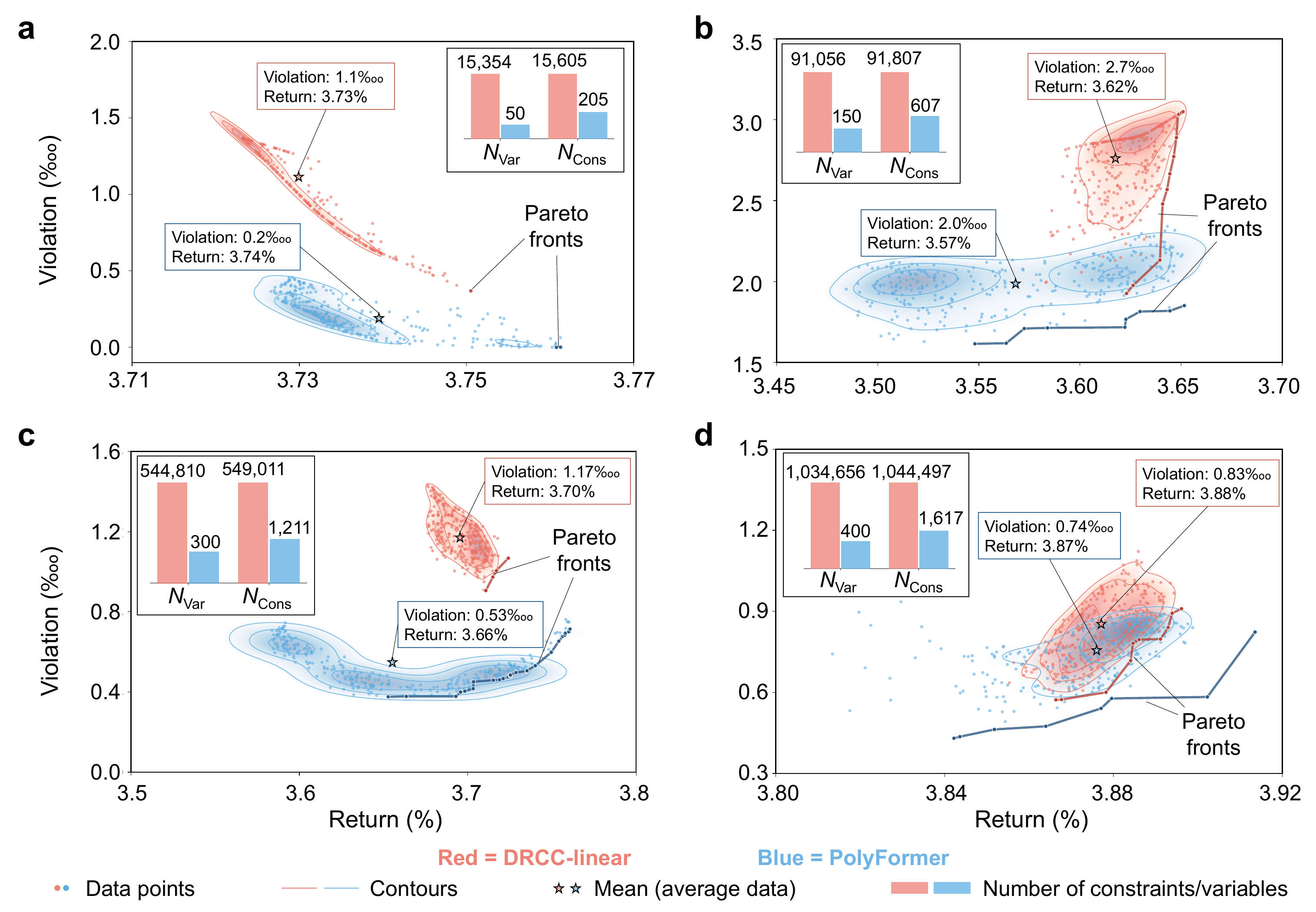}
\caption{\textbf{Benchmark results of DRCC portfolio optimization across four case settings. a,} 50 assets, 2 groups, 150 samples. \textbf{b,} 150 assets, 3 groups, 300 samples. \textbf{c,} 300 assets, 5 groups, 900 samples. \textbf{d,} 400 assets, 8 groups, 1280 samples.
Each scatter plot shows the return rate versus average constraint violation for 300 portfolio strategies solved using the DRCC-linear and PolyFormer models. Data points represent the average constraint violation and actual return of a portfolio strategy evaluated on a test dataset of 100 uncertain return samples (see \textcolor{blue}{Supplementary Note 5}). Here, constraint violation denotes the empirical average violation of the minimum-return constraint over the test samples and is distinct from PolyFormer's geometric feasibility error. The strategies are derived from varying parameters (risk level, Wasserstein ball radius, maximum group investment cap, and minimum acceptable return per group), as described in the ``Apply PolyFormer to DRCC portfolio optimization'' section in the Methods. Contours, averages, and Pareto fronts are annotated in the plots. Solutions with higher returns and lower errors (i.e., located towards the bottom-right) are considered better. The bar charts in each subfigure compare the number of variables and constraints for the two methods in each case. PolyFormer significantly reduces the number of variables (by 99.67\% to 99.99\%) and constraints (by 98.69\% to 99.85\%) compared to DRCC-linear in all cases, leading to a substantial decrease in computational complexity. 
}\label{fig:drcc}
\end{figure}

As a benchmark, we adopt the well-known Wasserstein-metric-based linear reformulation of DRCCs \cite{ordoudis_energy_2021}, hereafter referred to as ``DRCC-linear.'' The results obtained from PolyFormer's compact representation of DRCCs are compared with those obtained from the DRCC-linear model across four scenarios with different dimensionality, as shown in Fig. \ref{fig:drcc}. In terms of model complexity, PolyFormer dramatically reduces the problem size from up to 1,044,497 constraints and 1,034,656 variables to 1,617 constraints and 400 variables, corresponding to reductions of over 99.8\% and 99.9\%, respectively. This contraction translates into significant computational savings: it reduces solution time from 513 s to 0.725 s (708$\times$ faster) and memory consumption from 938 MB to 1.20 MB (99.87\% reduction). Across all four scenarios, PolyFormer attains smaller mean constraint violations and comparable mean returns. Its solution set also extends beyond the DRCC-linear Pareto frontier in the evaluated configurations. Combined with the 708-fold speedup, these results establish a favourable risk--return--computation trade-off.

Overall, for optimization under uncertainty, PolyFormer compresses the large set of auxiliary variables and constraints into a small set of constraints defined only over the original decision variables. This compression enables efficient exploration of risk-control parameters across alternative portfolio strategies. In practical deployments, the decision-maker can change risk-control parameters (e.g., confidence levels, conservatism factors) to generate multiple candidate portfolio strategies and select the most desirable trade-off between risk and return. With state-of-the-art methods such as DRCC-linear, each strategy computation can be prohibitively expensive. By contrast, PolyFormer enables rapid instantiation across parameter values and hence efficient generation and evaluation of many strategies. This capability markedly reduces the computational burden of large-scale optimization under uncertainty. Although PolyFormer is here evaluated on the portfolio optimization problem in a financial context, optimization under uncertainty is pervasive across fields like transportation \cite{WANG2024102923}, control \cite{Coulson2022Distributionally}, robotics \cite{Wu2024Data}, supply chain management \cite{GAO2024106803}, and machine learning \cite{LIN2024106755}. The same geometric compression principle offers a route to accelerating uncertainty-aware decision-making in these domains.

\section{Discussion}
This work establishes PolyFormer as a PIML framework for learning compact, solver-compatible representations of complex constraint geometry. Rather than predicting solutions, PolyFormer learns to simplify the optimization problem itself by replacing high-dimensional constraints with reusable polytopic representations. Its training accommodates mixed continuous and discrete variables, while neural parameterization enables rapid adaptation to changing operating conditions.

Across resource aggregation, two-layer power-system optimization, and portfolio optimization, PolyFormer substantially reduces model size and achieves online solver speedups of up to 6,400-fold. It eliminates up to 99.99\% of variables and 99.85\% of constraints while maintaining small feasibility and objective errors. These consistent gains demonstrate a unified constraint-simplification principle rather than an application-specific acceleration technique.

Looking forward, PolyFormer aligns with broader efforts to integrate domain knowledge and machine learning for solving various constrained optimization problems. Because PolyFormer represents each learned region as a convex polytope, nonconvex or disconnected regions may be convexified, and small errors over sampled directions do not provide automatic inclusion certificates. The framework is inherently extensible, with clear pathways for incorporating richer prior information, more expressive learning architectures, and tighter integration with modern optimization pipelines. Methodological extensions, including adaptive sampling methods for nonconvex regions and enhanced uncertainty handling, may further strengthen its robustness, particularly in safety-critical applications.

The implementation of PolyFormer is publicly available (see ``Code availability''), supporting reproduction and extension. More broadly, PolyFormer bridges realistic modelling and scalable decision-making by transforming complex constraint geometry into compact computational representations. This capability opens a route to rapid prescriptive optimization in complex systems across scientific, engineering, and socio-economic systems.

\section{Methods}
\subsection{Error metrics}
The Hausdorff distance is commonly used to quantify discrepancies between nonempty compact sets \cite{jungeblut_complexity_2024}. However, evaluating this worst-case metric can be computationally demanding, depending on how the sets are represented. We therefore construct tractable discrepancy measures from support and projection problems evaluated along sampled directions.

For a direction \(\mathbf v\in\mathbb R^n\setminus\{\mathbf0\}\), we define the directional feasibility and optimality errors as follows:
\begin{equation}\label{eq:dir_err_feas}
    {e_{{\text{feas}}}}({\mathbf{v}}) = \min\limits_{{\mathbf{z}} \in \Omega } {\left\| {{\mathbf{z}} - {\mathbf{x'}}} \right\|^2},{\text{s}}{\text{.t}}{\text{.,}}{\mathbf{x'}} = \mathop {\arg \max }\limits_{{\mathbf{x}} \in {{\mathcal P}}({\mathbf{A}},{\mathbf{b}})} {{\mathbf{v}}^{{\top}}}{\mathbf{x}},
\end{equation}
\begin{equation}\label{eq:dir_err_opt}
{e_{{\text{opt}}}}({\mathbf{v}}) = \min \limits_{{\mathbf{x}} \in {{\mathcal P}}({\mathbf{A}},{\mathbf{b}})} {\left\| {{\mathbf{x}} - {\mathbf{z'}}} \right\|^2},{\text{s}}{\text{.t}}{\text{.,}}{\mathbf{z'}} = \mathop {\arg \max }\limits_{{\mathbf{z}} \in \Omega } {{\mathbf{v}}^{{\top}}}{\mathbf{z}},
\end{equation}
Here, $e_{\text{feas}}({\mathbf{v}})$ is the squared distance from the support point $\mathbf x'$ of $\mathcal P$ to $\Omega$, whereas $e_{\text{opt}}({\mathbf{v}})$ is the squared distance from the support point $\mathbf z'$ of $\Omega$ to $\mathcal P$. For nonempty compact convex sets, the maximum of the two directional-error suprema over all directions equals the squared Hausdorff distance, $d_{\mathrm H}^2(\mathcal P,\Omega)$ \cite{jungeblut_complexity_2024}.

In \eqref{eq:dir_err_feas} and \eqref{eq:dir_err_opt}, the support subproblems identify points of \(\mathcal P\) and \(\Omega\) exposed along \(\mathbf v\), respectively. For nonconvex \(\Omega\), its support function coincides with that of its convex hull, so $e_{\mathrm{opt}}(\mathbf v)$ measures discrepancies from $\operatorname{conv}(\Omega)$. By contrast, the projection in $e_{\mathrm{feas}}(\mathbf v)$ detects nonconvex boundary violations encountered at the exposed points of $\mathcal P$. Together, the two metrics provide complementary training signals for a convex-hull-referenced approximation refined by detected nonconvex boundary violations.

For tractable training, we aggregate the directional errors by taking expectations over directions \(\mathbf v\) drawn from the standard $n$-dimensional Gaussian distribution:
\begin{equation}\label{eq:feas_opt_err_from_dir_err}
    e_{\text{feas}} = \mathbb{E}_{\mathbf{v}\sim \mathcal{N}(0,I_n)} e_{\text{feas}}(\mathbf{v}),\qquad e_{\text{opt}} = \mathbb{E}_{\mathbf{v}\sim \mathcal{N}(0,I_n)} e_{\text{opt}}(\mathbf{v}).
\end{equation}
Because the Gaussian distribution is rotationally invariant, the direction of a sample \(\mathbf v\sim\mathcal N(0,I_n)\) is uniformly distributed. At each iteration, PolyFormer samples one direction and evaluates the corresponding directional errors. Repeated sampling provides stochastic estimates of the expected weighted discrepancy.

For each sampled direction, evaluating the two directional errors requires only four support or projection problems over $\mathcal P$ and the original region $\Omega$, with linear or quadratic objectives. The training cost is therefore governed by the complexity of optimizing over the original region, while the number of solver calls grows linearly with the number of training iterations. Unlike analytical approximation methods based on iterative vertex--facet refinement or homothetic aggregation \cite{jones_polytopic_2010,wang_aggregate_2021}, PolyFormer introduces no additional nonlinear or combinatorial growth in optimization complexity beyond that inherited from the original region. Gurobi \cite{gurobi}, IPOPT \cite{wachter_implementation_2006}, BARON \cite{sahinidis_baron_1996}, or a comparable solver is selected according to the subproblem structure. Once trained, the learned polytope replaces the original constraints in subsequent optimization, yielding the online computational savings reported in the Results.


\subsection{Loss function and gradients}
This section derives an explicit loss function for training $\mathbf A$ and $\mathbf b$ based on the directional error calculations. Taking the directional feasibility error in \eqref{eq:dir_err_feas} as an example, suppose the lower-level linear program attains its optimum at $\mathbf{x}'$, and let $\mathbf{z}^\star$ denote the optimal solution of the upper-level projection problem. The directional feasibility error $e_{\text{feas}}({\mathbf{v}})$ is determined by $\mathbf{x}'$ and $\mathbf{z}^\star$, both of which depend implicitly on $\mathbf A$ and $\mathbf b$. Therefore, computing the gradient of $e_{\text{feas}}({\mathbf{v}})$ with respect to $\mathbf A$ and $\mathbf b$ requires applying implicit differentiation, which leads to nontrivial derivative expressions. Although such gradients can be obtained automatically using tools such as CVXPYLayers \cite{cvxpylayers2019}, this approach requires that all ${g_j}({\mathbf{x}},{\mathbf{y}})$, $j = 1,2, \cdots, m$, be convex and that both $\mathbf{x}$ and $\mathbf{y}$ are continuous variables, limiting its generality. Moreover, computing implicit differentiation involves large matrix inversions, which increase computational cost and may affect numerical stability.

Rather than differentiating the directional error exactly, PolyFormer constructs a surrogate loss and derives explicit approximate gradients for updating the polytope. Let $J$ index the constraints active at the support solution $\mathbf x'$. When the support solution is a nondegenerate vertex, the corresponding full-rank active constraints locally determine $\mathbf x'$. At each iteration, the selected hyperplanes are updated towards the projected boundary point $\mathbf z^\star\in\Omega$. For degenerate vertices or optimal faces, the solver's deterministic tie-breaking rule selects a support point and an associated active set for constructing the local surrogate. Assuming that each row of $\mathbf A$ is unit-normalized, the feasibility surrogate is
\begin{equation}\label{eq:feas_loss}
L_{\text{feas}}({\mathbf{A}},{\mathbf{b}};{\mathbf{v}}) = \|{\mathbf{A}}_J{\mathbf{z}}^\star - \mathbf{b}_J\|^2.
\end{equation}
For fixed $\mathbf x'$, $\mathbf z^\star$, and active set $J$, geometric analysis gives ${L_{{\text{feas}}}}(\mathbf{A}, \mathbf{b}; \mathbf{v}) = {e}_{\text{feas}}(\mathbf{v})\sum_{j \in J}\cos^2\phi_j$, where $\phi_j$ is the angle between $\mathbf z^\star-\mathbf x'$ and the normal of active hyperplane $j$. Under continuous directional sampling, the event that $\cos\phi_j=0$ for every $j\in J$ has probability zero, so the multiplicative factor is positive almost surely. Therefore, $L_{\text{feas}}$ and $e_{\text{feas}}(\mathbf v)$ have the same zero condition almost surely for the current support and projection points. Thus, the surrogate translates the current geometric discrepancy into a local correction that moves the active hyperplanes towards $\mathbf z^\star$.

The points $\mathbf x'$ and $\mathbf z^\star$ and the active set $J$ vary implicitly with $\mathbf A$ and $\mathbf b$. We apply a stop-gradient approximation by treating their current values as fixed during each update and ignoring this implicit dependence. Under this approximation, differentiating \eqref{eq:feas_loss} with respect to each row $\mathbf A_j$ and offset $b_j$ gives the following explicit approximate gradients:
\begin{equation}\label{eq:feas_loss_grad_A}
{\nabla _{{{\mathbf{A}}_j}}}{L_{{\text{feas}}}}({\mathbf{A}},{\mathbf{b}};{\mathbf{v}}) = \left\{ \begin{array}{l}
2({\mathbf{A}}_j{{\mathbf{z}}^ \star } - {{b}_j}){{\mathbf{z}}^{ \star \top }},j \in J\\
\qquad\;{{\mathbf{0}}^ \top }\qquad\;,{\text{otherwise}}
\end{array} \right.,
\end{equation}
\begin{equation}\label{eq:feas_loss_grad_b}
{\nabla _{{{b}_j}}}{L_{{\text{feas}}}}({\mathbf{A}},{\mathbf{b}};{\mathbf{v}}) = \left\{ \begin{array}{l}   
-2({\mathbf{A}}_j{{\mathbf{z}}^ \star } - {{b}_j}),j \in J\\
\qquad\;0\qquad\;,{\text{otherwise}}
\end{array} \right..
\end{equation}
Equations \eqref{eq:feas_loss_grad_A} and \eqref{eq:feas_loss_grad_b} update only the active hyperplanes, reducing their residuals at the current target point $\mathbf z^\star$. This update moves the selected local boundary towards the projected boundary point. At the next iteration, the support and projection problems are solved again, allowing the target points and active sets to adapt to the updated polytope.

The same construction applies to the optimality error $e_{\text{opt}}({\mathbf{v}})$. Suppose the projection problem in \eqref{eq:dir_err_opt} attains an optimum at $\mathbf{x}^\star$, with active constraints indexed by $K$, such that $\mathbf{A}_K\mathbf{x}^\star=\mathbf{b}_K$. Let $\mathbf z'$ denote the corresponding support point of $\Omega$. During differentiation, the current values of $\mathbf x^\star$, $\mathbf z'$, and $K$ are likewise treated as fixed. The optimality surrogate is
\begin{equation}
L_{\text{opt}}({\mathbf{A}},{\mathbf{b}};{\mathbf{v}}) = \|{\mathbf{A}}_K{\mathbf{z}}' - \mathbf{b}_K\|^2.
\end{equation}
The optimality construction is geometrically symmetric to the feasibility construction, so its surrogate and directional error likewise share the same zero condition almost surely.
Applying the same approximation to this surrogate gives explicit approximate gradients with respect to $\mathbf A_k$ and $b_k$:
\begin{equation}\label{eq:opt_loss_grad_A}
{\nabla _{{{\mathbf{A}}_k}}}{L_{{\text{opt}}}}({\mathbf{A}},{\mathbf{b}};{\mathbf{v}}) = \left\{ \begin{array}{l}
2({\mathbf{A}}_k{{\mathbf{z}}' } - {{b}_k}){{\mathbf{z}}'^{ \top }},k \in K\\
\qquad\;{{\mathbf{0}}^ \top }\qquad\;,{\text{otherwise}}
\end{array} \right.,
\end{equation}
\begin{equation}\label{eq:opt_loss_grad_b}
{\nabla _{{{b}_k}}}{L_{{\text{opt}}}}({\mathbf{A}},{\mathbf{b}};{\mathbf{v}}) = \left\{ \begin{array}{l}
-2({\mathbf{A}}_k{{\mathbf{z}}' } - {{b}_k}),k \in K\\
\qquad\;0\qquad\;,{\text{otherwise}}
\end{array} \right..
\end{equation}
Weighting the two surrogate losses by $\lambda$ yields the total PolyFormer loss:
\begin{equation}\label{eq:total_loss}
\begin{aligned}
L_{\text{total}}({\mathbf{A}},{\mathbf{b}};{\mathbf{v}}) &= \lambda {L_{{\text{feas}}}}({\mathbf{A}},{\mathbf{b}};{\mathbf{v}}) + (1 - \lambda ){L_{{\text{opt}}}}({\mathbf{A}},{\mathbf{b}};{\mathbf{v}})\\
&= \lambda \|{\mathbf{A}}_J{\mathbf{z}}^\star - \mathbf{b}_J\|^2 + (1 - \lambda )\|{\mathbf{A}}_K{\mathbf{z}}' - \mathbf{b}_K\|^2.
\end{aligned}
\end{equation}
The corresponding gradients are
\begin{equation}\label{eq:total_loss_grad_A}
{\nabla _{{{\mathbf{A}}}}}{L_{{\text{total}}}}({\mathbf{A}},{\mathbf{b}};{\mathbf{v}}) = \lambda {\nabla _{{{\mathbf{A}}}}}{L_{{\text{feas}}}}({\mathbf{A}},{\mathbf{b}};{\mathbf{v}}) + (1 - \lambda ){\nabla _{{{\mathbf{A}}}}}{L_{{\text{opt}}}}({\mathbf{A}},{\mathbf{b}};{\mathbf{v}}),
\end{equation}
\begin{equation}\label{eq:total_loss_grad_b}
{\nabla _{{\mathbf{b}}}}{L_{{\text{total}}}}({\mathbf{A}},{\mathbf{b}};{\mathbf{v}}) = \lambda {\nabla _{{\mathbf{b}}}}{L_{{\text{feas}}}}({\mathbf{A}},{\mathbf{b}};{\mathbf{v}}) + (1 - \lambda ){\nabla _{{\mathbf{b}}}}{L_{{\text{opt}}}}({\mathbf{A}},{\mathbf{b}};{\mathbf{v}}).   
\end{equation}

Equations \eqref{eq:total_loss_grad_A} and \eqref{eq:total_loss_grad_b} give the approximate gradients with respect to $\mathbf A$ and $\mathbf b$ at each iteration. Automatic differentiation computes the same approximate gradients when the support points, projection points, and active-set selections are detached from the computational graph. Different sampled directions activate different index sets $J$ and $K$, so successive boundary-directed updates progressively reshape the polytope.

The surrogate loss extends to parameterized regions $\Omega(\boldsymbol\theta)$ by replacing $\mathbf A$ and $\mathbf b$ with differentiable mappings $\mathbf A(\boldsymbol\theta;\boldsymbol w_{\mathbf A})$ and $\mathbf b(\boldsymbol\theta;\boldsymbol w_{\mathbf b})$. PolyFormer typically implements these mappings using neural networks, consistent with the parameterization introduced in the Overview. During training, $\boldsymbol\theta$ is a sampled input, whereas $\boldsymbol w_{\mathbf A}$ and $\boldsymbol w_{\mathbf b}$ are the trainable network parameters. The approximate gradients with respect to $\mathbf A$ and $\mathbf b$ are propagated through the two networks by the chain rule to obtain $\nabla_{\boldsymbol w_{\mathbf A}}L_{\mathrm{total}}$ and $\nabla_{\boldsymbol w_{\mathbf b}}L_{\mathrm{total}}$. Our implementation uses PyTorch \cite{paszke_pytorch_2019}.

\subsection{Implementation remarks}
Here, we provide three important remarks to facilitate a smooth implementation of PolyFormer.
\begin{remark}[Initialization]\label{rem:init}
The initial approximate polytope must be nonempty and bounded. We initialize the rows of \(\mathbf{A}\) randomly or from domain knowledge (see \textcolor{blue}{Supplementary Notes~2--5} for the case-specific strategies). Randomly generated rows are accepted only when they pass a boundedness test; otherwise, they are resampled. Given an accepted \(\mathbf{A}\), we initialize \(\mathbf{b}\) by solving \(b_j = \max_{\mathbf{x}\in \Omega} \mathbf{A}_j \mathbf{x}\), \(1\le j \le M\). Finite solutions to these support problems yield a nonempty and bounded outer approximation of \(\Omega\), for which \(e_{\text{opt}}=0\). For parameterized regions, we perform this initialization at the mean value of \(\boldsymbol\theta\) to provide a geometrically reasonable warm start. In the neural networks that fit \(\mathbf{A}(\boldsymbol\theta)\) and \(\mathbf{b}(\boldsymbol\theta)\), all coefficients are then initialized to zero except for the bias terms, which are initialized with the resulting values of \(\mathbf{A}\) and \(\mathbf{b}\).
\end{remark}
\begin{remark}[Normalization]\label{rem:normal}
    As mentioned in the derivation of the loss function and gradients \eqref{eq:feas_loss}-\eqref{eq:total_loss_grad_b}, the row vectors of \(\mathbf{A}\) need to be normalized throughout the entire training process. This can be achieved by dividing each row vector of \(\mathbf{A}\) and the corresponding element of \(\mathbf{b}\) by the norm of that row vector after each update of \(\mathbf{A}\) and \(\mathbf{b}\). We also recommend normalizing each dimension of the original variable \(\mathbf{x}\) and the parameter \(\boldsymbol \theta\) (if there are any), e.g., constraining each dimension of \(\mathbf{x}\) and \(\boldsymbol \theta\) within \([0,1]\), which can improve the training stability and convergence speed.
\end{remark}
\begin{remark}[Setting $\lambda$]\label{rem:lambda}
The weight coefficient $\lambda$ controls the emphasis on feasibility and optimality during training. Typically, we recommend starting with a balanced value, e.g., $\lambda = 0.5$, so that the feasibility error and optimality error can compete with each other to identify the shape that best matches the original region $\Omega$. After training for enough iterations, $\lambda$ can be adjusted based on the specific requirements of the practical scenario. For example, in optimization scenarios that prioritize feasibility, increasing $\lambda$ places greater emphasis on reducing the feasibility error.
\end{remark}

\subsection{Typical geometries}

To help better understand the training procedure of PolyFormer and how its outputs adapt to the varying parameters, we provide three illustrative examples, i.e., polygons, ellipses, and nonconvex regions, that offer an intuitive introduction to PolyFormer. In addition, to demonstrate PolyFormer's accuracy and scalability, we also implemented PolyFormer on two benchmark cases with known global optima (hypercubes and balls). The numerical settings for all cases in this section are provided in \textcolor{blue}{Supplementary Note 2}.

In the three 2D cases, we use PolyFormer to obtain a quadrilateral approximation of an octagon, a hexagonal approximation of an ellipse, and a hexagonal approximation of a nonconvex region (a portion of a disk that does not intersect another disk). As shown in \textcolor{blue}{Extended Data Fig. 1}a, the approximating polygons evolve over successive iterations, exhibiting a trade-off between feasibility and optimality errors, and converge to shapes that closely match the original regions in all three cases. When introducing varying parameters \(\boldsymbol{\theta}\), the underlying feasible set becomes parameter-dependent. \textcolor{blue}{Extended Data Fig. 1}b-d shows that the true feasible region deforms smoothly as \(\boldsymbol{\theta}\) varies, and, correspondingly, the parameterized PolyFormer produces polytopes that adapt coherently to these changes. Notably, in the nonconvex case, the final polygon fits the convex hull of the original nonconvex region, which has been stated in the ``Error metrics'' section in the Methods. While this reflects a limitation of the current PolyFormer, convex hulls are adequate in many applications \cite{Yang2024Convex,de_rosa_explicit_2024, adams_hierarchy_2005}. In applications where tighter feasibility guarantees are required for nonconvex sets, extending the sampling strategy to capture a convex inner approximation, e.g., via fixed-point rays, offers a natural direction for future work.

In the hypercube case, the target feasible set is an $n$-dimensional hypercube represented exactly by $M=2n$ axis-aligned hyperplanes, so the globally optimal approximation error is zero. \textcolor{blue}{Extended Data Fig. 2}a reports the initial error across dimensions and the number of training steps needed for the total error to drop below \(10^{-5}\). Across the tested range \(2\leq n\leq200\), PolyFormer reaches this threshold in every case, although both the initial error and the required number of training steps show an overall increase with dimension.

In the ball case, the target feasible set is an \(n\)-dimensional unit ball, and the approximate polytope again uses \(M = 2n\) hyperplanes. For $\lambda=0.5$, the globally minimal total error is \(1 - \frac{2\sqrt{n}}{n+1}\), which provides an analytical benchmark. \textcolor{blue}{Extended Data Fig. 2}b compares the initial and converged errors with this analytical reference and shows that the converged errors closely track it across the tested dimensions. PolyFormer achieves an error reduction exceeding 99.4\% across all tested dimensions. Notably, in higher-dimensional regimes ($n \ge 30$), the error reduction consistently surpasses 99.9\%.

\subsection{Apply PolyFormer to resource aggregation}
We study the aggregation of feasible power regions for EVs, HPs, and BSSs, a task that arises commonly in smart grid applications \cite{chen_aggregate_2020,wang_aggregate_2021}. Let \(\mathcal{I}_{\text{EV}}\), \(\mathcal{I}_{\text{BSS}}\) and \(\mathcal{I}_{\text{HP}}\) denote the corresponding resource sets, indexed by \(i\). Let \(\mathbf{p}_i = [p_{i,1},\ldots,p_{i,T}]^{\top}\in\mathbb{R}^T\) denote the power trajectory of resource \(i\) (with load treated as positive), where \(T=24\) represents a 24-hour horizon.

Each resource is subject to specific physical and technological constraints that define an individual feasible region \(\Omega_i\) for its power trajectory: \(\mathbf{p}_i \in \Omega_i, \forall i \in \mathcal{I}_{\text{EV}} \cup \mathcal{I}_{\text{BSS}} \cup \mathcal{I}_{\text{HP}}\). The exact definitions of \(\Omega_i\) for each resource type are provided in \textcolor{blue}{Supplementary Note 3}. We aim to characterize the feasible region of the aggregated power trajectory \(\mathbf{P} = [P_1,\ldots,P_T]^{\top}\in \mathbb{R}^T\), defined by the Minkowski sum \cite{tiwary_hardness_2008,Schneider_2013}:
\begin{equation}
\Omega = \left\{ \mathbf{P} \in \mathbb{R}^T \,\middle|\,
\begin{aligned}
P_t &= \sum_{i\in \mathcal{I}_{\text{EV}} \cup \mathcal{I}_{\text{BSS}} \cup \mathcal{I}_{\text{HP}}} p_{i,t}, \quad \forall t\in\{1,2,\cdots,T\}\\
\mathbf{p}_i &\in \Omega_i, \quad \forall i\in \mathcal{I}_{\text{EV}} \cup \mathcal{I}_{\text{BSS}} \cup \mathcal{I}_{\text{HP}}
\end{aligned}
\right\}.
\label{eq:agg_region}
\end{equation}

The aggregated power trajectory \(\mathbf{P}\) corresponds to the decision variable \(\mathbf{x}\) in Equation \eqref{eq:orig_region} with dimension \(n=T\), and the collection of individual trajectories \(\mathbf{p}_i, \forall i \in \mathcal{I}_{\text{EV}} \cup \mathcal{I}_{\text{BSS}} \cup \mathcal{I}_{\text{HP}}\) corresponds to the auxiliary variable \(\mathbf{y}\) with dimension \(n' = T(|\mathcal{I}_{\text{EV}}| + |\mathcal{I}_{\text{BSS}}| + |\mathcal{I}_{\text{HP}}|)\). 

We apply PolyFormer to construct a polytopic approximation of \(\Omega\), using \(M=4T\) hyperplanes (rows of \(\mathbf{A}\)). This case does not involve varying parameters, so we directly optimize \(\mathbf{A}\) and \(\mathbf{b}\) following the training pipeline in Fig. \ref{fig:framework}b. The three-phase training schedule prioritizes feasibility in its final phase and produces a near-inner approximation over the evaluated directions (see the caption of Fig. \ref{fig:aggregation}). Detailed numerical settings are provided in \textcolor{blue}{Supplementary Note 3}.

\subsection{Apply PolyFormer to the two-layer power system optimization}
In the two-layer power system optimization problem, both levels are governed by nonlinear power-flow constraints. Specifically, the upper-level transmission network is modelled via the standard AC power-flow equations \cite{Tinney1967Power}, while the lower-level distribution network adopts the DistFlow formulation \cite{Baran1989Network} (see \textcolor{blue}{Supplementary Note 4} for details). Within each distribution network, a subset of nodes can flexibly adjust active and reactive power injections within prescribed bounds through demand-response programs or controllable resources (e.g., distributed generation and energy storage) \cite{jackson_building_2021}. The combination of these flexibility envelopes and the network operating constraints induces a feasible set for the active-reactive power exchange at the transmission-distribution (T-D) interface. In practice, distribution networks are typically operated in a radial topology \cite{Deka2024Learning}, with the T-D interface corresponding to the root node of the distribution network (physically, a substation). Electrical consistency requires that the active power, reactive power, and voltage magnitude (per unit) at the transmission-side interconnection node coincide with the corresponding variables at the distribution root node. Variations in the root-node voltage propagate along the feeder and may push downstream nodal voltages beyond their allowable limits, thereby reshaping the admissible active-reactive power region at the root node. Consequently, the root-node feasible region in the \((P,Q)\) plane is inherently voltage-dependent. We denote this parametric feasible set by \((P,Q)\in \Omega(V)\), where \(P\), \(Q\), and \(V\) represent the root-bus active power, reactive power, and voltage magnitude, respectively.

We employ PolyFormer to approximate the voltage-parameterized feasible region \(\Omega(V)\) of each distribution network. For a fixed value of $V$, the approximation is a polytope described by linear inequalities in the root-node power injections $(P,Q)$. Accordingly, PolyFormer operates in a two-dimensional decision space (\(n=2\)). We fix the number of hyperplanes by setting the row dimension of \(\mathbf{A}\) to \(M=8\). In this setting, the sole varying parameter is the voltage magnitude \(V\); hence, the neural networks used to parameterize \(\mathbf{A}(\boldsymbol{\theta})\) and \(\mathbf{b}(\boldsymbol{\theta})\) take a one-dimensional input.\footnote{We use root-node voltage as the sole varying parameter to isolate voltage-dependent adaptation in this benchmark. In deployments with broader operating variability, exogenous operating conditions, such as load levels, renewable generation, and equipment availability, can be added as neural-network inputs.} The network architecture consists of a single fully connected hidden layer with 128 neurons and ReLU activation placed between the input and output layers. During training, we uniformly sample \(V\) over its admissible interval, i.e., $[0.95, 1.05]$ p.u., to generate a diverse set of operating conditions. The detailed numerical configuration is provided in \textcolor{blue}{Supplementary Note 4}. In the present benchmark, the voltage parameter is handled through a one-shot instantiation procedure. We first obtain initial interface-voltage estimates $\widehat V_i$ from baseline distribution- and transmission-network power-flow calculations. Each trained PolyFormer is then evaluated at $\widehat V_i$, and the resulting numerical coefficients are held fixed in the simplified transmission-network optimization as $\mathbf A_i(\widehat V_i)[P_i,Q_i]^\top\leq\mathbf b_i(\widehat V_i)$. Thus, the neural networks generate the polytopic constraints before optimization and are not explicitly embedded in the online optimization model.

\subsection{Apply PolyFormer to DRCC portfolio optimization}
The DRCC portfolio optimization is formulated as follows:
\begin{subequations}
    \begin{align}
    & \max && \sum\nolimits_{i = 1}^N {{{\overline r }_i}} {x_i} \label{eq:portfolio_obj}\\
    & \text{s.t.} && \inf\nolimits_{\mathbb{P} \in \mathcal{Q}_g} \mathbb{P}(\sum\nolimits_{i \in {{{\mathcal N}}_g}} {{{\tilde r}_i}{x_i}}  \ge {R_g}\sum\nolimits_{i \in {{{\mathcal N}}_g}} {{x_i}} ) \ge 1 - \epsilon_g, \quad g = 1, 2, \ldots, G, \label{eq:portfolio_drcc}\\
    & && \sum\nolimits_{i\in \mathcal N_g} x_i \le X^{\max}_g, \quad g = 1, 2, \ldots, G,\label{eq:portfolio_group}\\
    & && \sum\nolimits_{i=1}^{N} x_i \le 1, \label{eq:portfolio_total}
\end{align}
\end{subequations}
where the decision variable \(x_i \in \mathbb{R}_+\) represents the proportion of investment in asset \(i\); the uncertain variable \(\tilde{r}_i\) denotes the random return of asset \(i\), and \(\overline{r}_i\) denotes its expectation; $N$ is the total number of assets; $G$ is the total number of asset groups; \(\mathcal{N}_g\) is the set of assets in group \(g\); \(R_g\) is the minimum acceptable return for group \(g\); \(\epsilon_g\) is the risk level for group \(g\); \(X^{\max}_g\) is the maximum allowable investment proportion for group \(g\); and \(\mathcal{Q}_g\) represents the ambiguity set for asset group \(g\), defined using the Wasserstein distance with radius \(\rho_g\) and centred at the empirical distribution derived from a historical dataset of size \(K\) \cite{xie_distributionally_2021}.

The objective function \eqref{eq:portfolio_obj} maximizes the expected return of the portfolio. The DRCC \eqref{eq:portfolio_drcc} ensures that, for each asset group, the probability of the portfolio return falling below the specified minimum acceptable return does not exceed the given risk level $\epsilon_g$, considering the worst-case distribution within the ambiguity set. Constraint \eqref{eq:portfolio_group} enforces the group-wise investment caps, whereas constraint \eqref{eq:portfolio_total} limits the total investment. The challenge of solving this problem lies in the complexity of the DRCCs \eqref{eq:portfolio_drcc}, which involve infinite-dimensional optimization over probability distributions. We first apply an analytical reformulation to convert the DRCCs into a finite set of linear constraints \cite{ordoudis_energy_2021}, as detailed in \textcolor{blue}{Supplementary Note 5}. PolyFormer then approximates the joint feasible region defined by the reformulated DRCC and the group-wise investment cap with a small number of linear inequalities involving only the decision variables \(x_i\).

The variable dimension for the PolyFormer used to approximate the feasible region of asset group $g$ is $n = |\mathcal{N}_g|$, and the number of rows of the matrix $\mathbf{A}$ is set to $M = 4n+2$. There are four parameters for each asset group: the risk level $\epsilon_g$, the ambiguity-set radius $\rho_g$, the minimum acceptable return $R_g$, and the groupwise investment cap $X^{\max}_g$. Therefore, the input dimension for the neural networks that parameterize $\mathbf{A}(\boldsymbol \theta)$ and $\mathbf{b}(\boldsymbol \theta)$ is 4. 
We add a fully connected hidden layer with 128 neurons and ReLU activation between the input and output. During training, we randomly sample these four varying parameters within predefined ranges to generate diverse training scenarios. The specific parameter ranges and other numerical settings are detailed in \textcolor{blue}{Supplementary Note 5}. When the training is complete, the learned polytopes are directly embedded into the portfolio optimization problem to replace the group-level DRCCs \eqref{eq:portfolio_drcc} and investment-cap constraints \eqref{eq:portfolio_group}, while the total-investment constraint \eqref{eq:portfolio_total} remains explicit.



\backmatter




\bmhead{Acknowledgements}
This work is funded by the National Natural Science Foundation of China under Grant Numbers 52507091 and U24B2080; Smart Grid-National Science and Technology Major Project under Grant Number 2024ZD0801700; China Postdoctoral Science Foundation under Grant Number GZC20250335; State Key Laboratory of Power System Operation and Control under Grant Number SKLD25KZ04. We thank G. Hug for valuable discussions and guidance in the design of experiments.

\bmhead{Author contributions}
Y.W. conceived the idea. Y.W. and Y.G. developed the theory and methods. Y.W. implemented the algorithms. B.Z. acquired the data and provided the experimental conditions. Y.W., Y.G., and W.Q. conducted the numerical studies and analysis. Y.W., Y.G., and W.Q. prepared the figures. B.Z. and J.S. supervised the work and provided resources. All authors contributed to data analysis, discussions, and manuscript preparation.
\bmhead{Competing interests}
The authors declare no competing interests.

\bmhead{Data availability}
The datasets used to generate the reported numerical results and instructions for obtaining third-party benchmark data are available at \url{https://github.com/wenyl16/PolyFormer}.

\bmhead{Code availability}
The source code and scripts used to train and evaluate PolyFormer are available at \url{https://github.com/wenyl16/PolyFormer}.

\bibliography{ref}

\clearpage
\makeatletter
\renewcommand{\fnum@figure}{Extended Data Fig. 1}  
\begin{figure}[t]
\centering
\includegraphics[width=0.95\textwidth]{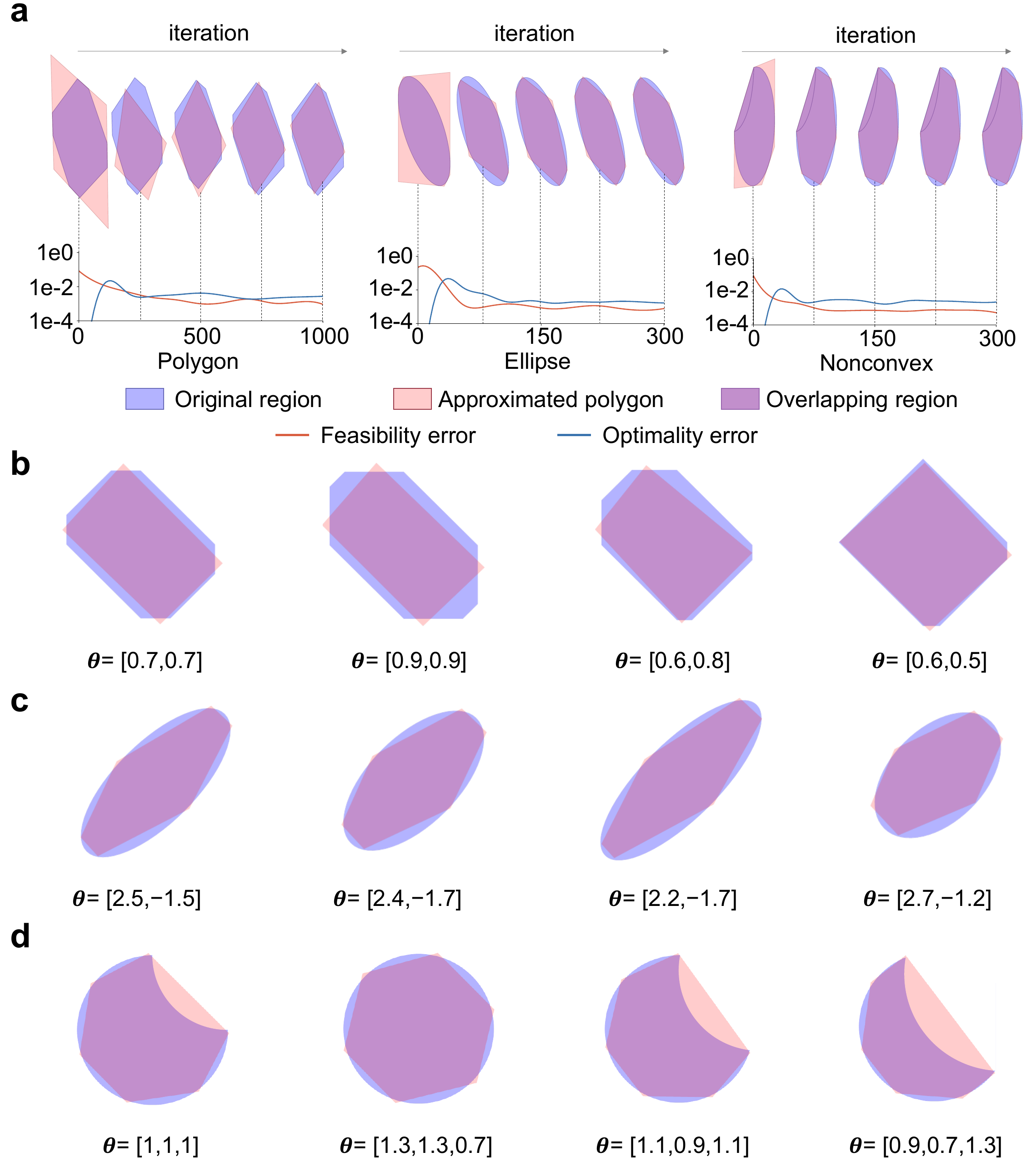}
\caption{\textbf{Illustrative results for three 2D cases. a,} Evolution of training errors and corresponding geometric shapes for polygonal, elliptical, and nonconvex regions, evaluated with all varying parameters fixed at their baseline values. As feasibility and optimality errors compete during training, the approximated regions progressively converge towards the original region.
\textbf{b--d,} PolyFormer fitting results under varying parameters for the polygon (b), ellipse (c), and nonconvex (d) cases. Changes in the parameters induce deformations of the original regions, which are consistently tracked by the approximating polygons.
} 
\label{fig:ExtendedData_2dshapes}
\end{figure}
\renewcommand{\fnum@figure}{\figurename~\thefigure}  
\makeatother

\clearpage
\makeatletter
\renewcommand{\fnum@figure}{Extended Data Fig. 2}  
\begin{figure}[t]
\centering
\includegraphics[width=0.85\textwidth]{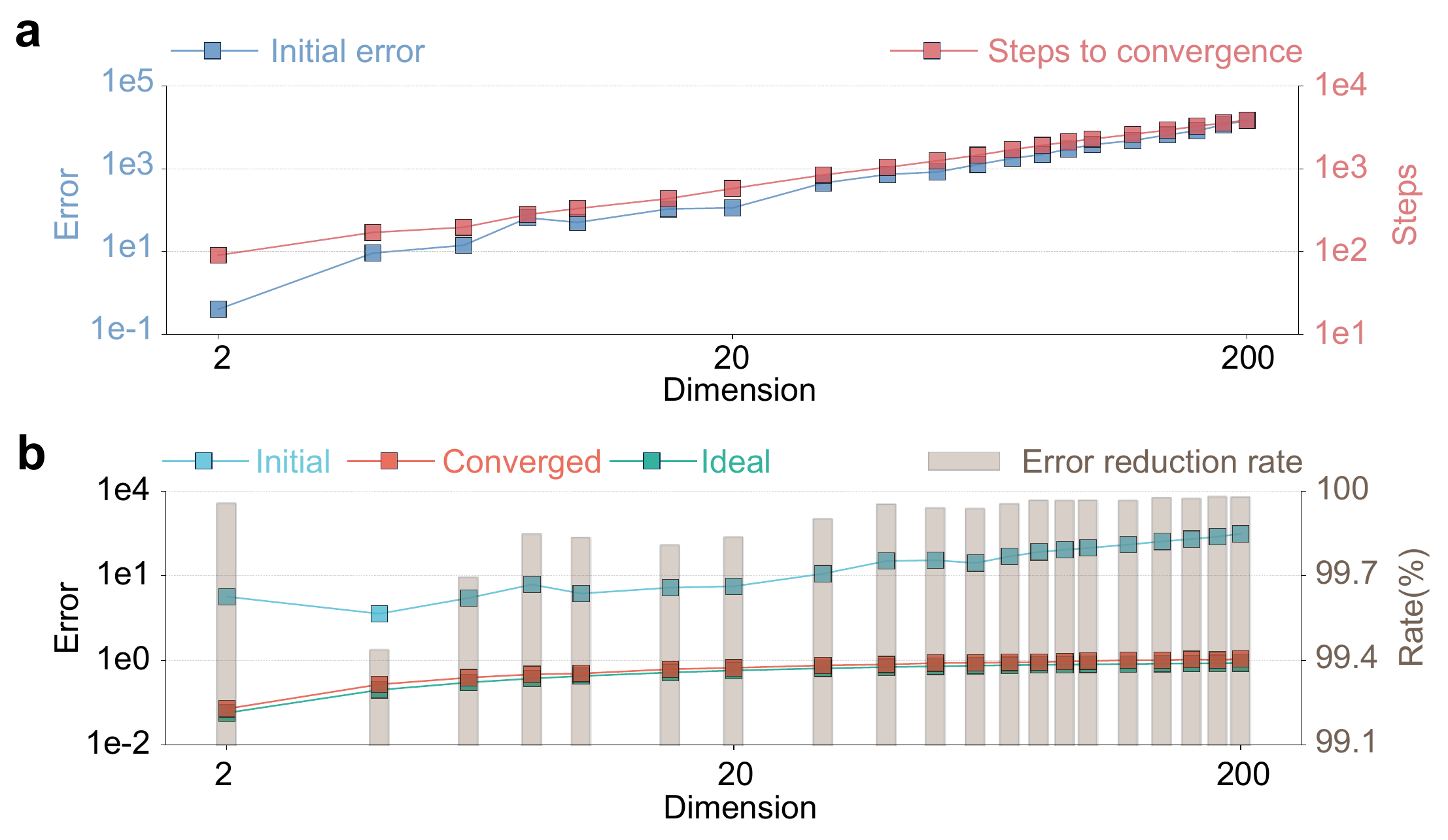}
\caption{\textbf{Benchmark results for two multi-dimensional cases. a,} Hypercube benchmark. For all tested dimensions, the final total error falls below \(10^{-5}\), indicating convergence. The initial total error (left) and the number of steps to convergence (right) both show overall upward trends with dimensionality. \textbf{b,} Unit-ball benchmark. The converged errors remain close to the globally minimal total errors across the tested dimensions. Line plots show the initial, converged, and analytical-reference total errors, while bar charts report the error reduction rate from initialization to convergence. Across all dimensions, the total error is reduced by more than 99.4\%, exceeding 99.9\% when \(n \geq 30\).
} 
\label{fig:ExtendedData_mdshapes}
\end{figure}
\renewcommand{\fnum@figure}{\figurename~\thefigure}  
\makeatother
\end{document}